\documentclass[runningheads]{llncs}
\usepackage[T1]{fontenc}
\usepackage{graphicx}

\usepackage{tabularray}

\usepackage{physics}
\usepackage{tabularx}
\usepackage{subcaption}
\usepackage{dingbat}
\usepackage{acro}
\usepackage{amsfonts}
\usepackage{amssymb}
\usepackage{soul}
\usepackage{cleveref}
\usepackage{stix}

\newcommand{\etal}{\textit{et~al.}}
\begin{document}
\title{Efficient Inter-Task Attention for Multitask Transformer Models}
%
%

\author{Christian Bohn\inst{1, 2,}\thanks{Corresponding author: \email{christian.bohn@uni-wuppertal.de}}\orcidID{0009-0006-5684-0318} \and
Thomas Kurbiel\inst{1}\orcidID{0009-0002-4249-9113} \and \\
Klaus Friedrichs\inst{1}\orcidID{0009-0005-9120-6796} \and
Hasan Tercan\inst{2}\orcidID{0000-0003-0080-6285} \and \\
Tobias Meisen\inst{2}\orcidID{0000-0002-1969-559X}}
\authorrunning{C. Bohn et al.}
%

\institute{APTIV \and
	University of Wuppertal, Germany}
%
\maketitle              
\begin{abstract}
In both Computer Vision and the wider Deep Learning field, the Transformer architecture is well-established as state-of-the-art for many applications. For Multitask Learning, however, where there may be many more queries necessary compared to single-task models, its Multi-Head-Attention often approaches the limits of what is computationally feasible considering practical hardware limitations. This is due to the fact that the size of the attention matrix scales quadratically with the number of tasks (assuming roughly equal numbers of queries for all tasks). 
As a solution, we propose our novel Deformable Inter-Task Self-Attention for Multitask models that enables the much more efficient aggregation of information across the feature maps from different tasks.
In our experiments on the NYUD-v2 and PASCAL-Context datasets, we demonstrate an order-of-magnitude reduction in both FLOPs count and inference latency. At the same time, we also achieve substantial improvements by up to 7.4\% in the individual tasks' prediction quality metrics. 

\end{abstract}

\section{Introduction} \label{introduction}
The two primary goals in Multitask Learning are to reduce the amount of compute resources required per task and to achieve better generalization by enabling the model to use information from one task to improve performance on another \cite{multitask_learning_caruana}. This is achieved by sharing parts of the model's learnable weights between tasks. Recently, with the rise of Transformer models \cite{attention_is_all_you_need}, this shared part of the architecture is often an attention module as it provides great adaptability to a wide variety of applications given enough training data \cite{Bhattacharjee_2022_CVPR,xyy2023DeMT,xu2022mtformer}. 
This adaptability does come with the significant shortcoming, however, that the compute resources required for the attention computation scale quadratically with the number of queries. 
Especially for multitask models where enabling attention across tasks is desirable, this scaling behavior becomes problematic. Assuming roughly comparable numbers of queries for the different tasks, the computational cost of the attention mechanism in such a model now scales quadratically with the number of tasks, defeating the purpose of Multitask Learning to reduce the amount of compute resources required per task. For Multitask Vision Transformer models that typically require a large number of queries that is proportional to the count of input pixels, this problem is even more relevant. 
Previous works approached this issue by either reducing the number of queries, as in \cite{Xu2022MultiTaskLW} or by merging task feature maps and thus attending to fewer values \cite{xu2022mtformer}. In \textit{Deformable Mixer Transformer (DeMT)} \cite{xyy2023DeMT}, Xu \etal accepted the computational overhead in favor of better prediction metrics compared to \cite{Xu2022MultiTaskLW} and \cite{xu2022mtformer}.
Inspired by the work on Deformable Attention by Zhu \etal \cite{zhu2021deformable}, we replace the global attention used in the DeMT model \cite{xyy2023DeMT} with our \textit{Deformable Inter-Task Self-Attention} that has been designed for the application with multiple task`s feature maps. In doing so, we allow the model to overcome the weaknesses of the aforementioned prior works by computing the attention across tasks much more efficiently with better scaling properties and in addition also achieve improved performance metrics. 
As also demonstrated in \cite{zhu2021deformable}, this improvement in prediction quality is likely due to poor convergence of learning global attention between every query-key pair. This suggests that for Multitask Models to be able to optimally leverage information from other tasks and improve the performance on one task, it is important that this information is efficiently accessible. 
We evaluate our method on multiple feature extraction backbones and different multitask Computer Vision (CV) datasets. \\
The main contributions of our work can be summarized as follows:
\begin{itemize}
    \item We introduce our novel, more efficient \textit{Deformable Inter-Task Self-Attention (ITSA)} mechanism. 
    \item We integrate this attention mechanism into the state-of-the-art DeMT model architecture. 
    \item We extensively evaluate the resulting model on both the NYUD-v2 and PASCAL-Context dataset, demonstrating both its reduced compute resource requirements and improved prediction quality. 
\end{itemize}

\section{Related Work}
\textbf{Vision Transformers.} 
After its introduction for image classification in \cite{dosovitskiy2021an}, the Vision Transformer has been established as the state-of-the-art Computer Vision model architecture in recent years, especially for more complex tasks. 
A common feature of Vision Transformers is that they assign queries to small patches of the input image. For a task like image classification, these patches can be somewhat larger, but for dense prediction tasks like segmentation, prediction quality suffers significantly if the patch size is chosen too large. While Dosovitskiy \etal in \cite{dosovitskiy2021an} were able to achieve good classification performance with patches of 16 by 16 pixels, anything larger than 8-by-8-patches in the Segmenter model \cite{strudel2021segmentertransformersemanticsegmentation} by Strudel \etal leads to quite poor results, and Xie \etal even used 4-by-4-patches in \cite{xie2021segformer}. This demonstrates the drawback of vision transformers that they need many queries, leading to expensive attention computation. For multiple tasks, this issue is exacerbated significantly. 

\textbf{Deformable Attention.} As mentioned in \Cref{introduction}, reducing the computational overhead associated with attention computation 
is a highly relevant problem. 
Significant efficiency improvements in the attention computation can be achieved by utilizing low-level optimizations such as \cite{dao2022flashattentionfastmemoryefficientexact,dao2023flashattention2fasterattentionbetter} that fully take the particular constraints of the used hardware into account. 
Works like \cite{liao2024viglinearcomplexityvisualsequence,csordás2024switchheadacceleratingtransformersmixtureofexperts} approach that problem by introducing gating into the attention computation so that only a relevant subset of the full Multihead Attention computation actually needs to be performed. 
For many applications, however, especially for inputs where some locality bias can be exploited, computing the full attention matrix between every pair of queries is not necessary and can be approximated well by Deformable Attention. This attention operation took inspiration from Deformable Convolution \cite{deformable_convolution} which generalizes the convolution operation by allowing sampling from arbitrary locations in the feature maps. These are defined by offsets computed with the convolution kernel. Applying the same idea to attention computation, \cite{zhu2021deformable} introduced Deformable Attention: Here, through learnable linear layers, a query feature defines both sampling offsets and attention weights: Those allow us to compute the attention output for that query by sampling the feature map at the offsets around a reference point for that query, weighting these samples by their respective attention weights, and finally summing the weighted samples. Usually the number of offsets chosen per query is small (16 is a common choice), thus making this operation much more efficient than traditional, full attention, where for every query, one needs to compute its interaction with every other. Despite that, \cite{zhu2021deformable} shows that deformable attention can actually outperform full attention while saving compute resources. 
In \cite{vision_transformer_w_def_attn}, Xia \etal argue that the lack of keys in the attention computation of \cite{zhu2021deformable} limits its representational power, thus making it unsuitable as a feature extraction backbone. Instead they modify the Deformable Attention computation to generate both a key and value from each sampled location and compute the attention as the cosine similarity between the query and the key for each sampled location. In our case however, since we are using a convolutional backbone and apply Deformable Attention on its extracted features, this is not a concern for us and we observe strong results following the simpler approach in \cite{zhu2021deformable}.
\\

\textbf{Multitask Learning.} 
Most Multitask Learning (MTL) works explore novel model architectures \cite{Bhattacharjee_2022_CVPR,xyy2023DeMT,bruggemann2020exploring,xu2022mtformer,Xu2022MultiTaskLW},  or focus on optimization methods for task balancing and loss weighting \cite{liu2021towards,Guo_2018_ECCV,PCGrad_paper,bohn2024taskweightinggradientprojection}.
While MTL has been applied to various learning domains such as natural language processing \cite{10.1145/3663363,zhang2023surveymultitasklearningnatural}, radar signal processing \cite{MTL_for_Radar}, and LiDAR point cloud analysis \cite{lang2024pointbasedapproachefficientlidar}, one of its primary applications - and the focus of this work - is Computer Vision. Common tasks in multitask CV include object detection \cite{KhattarHH21,UniT}, visual question answering \cite{UniT}, or 3D CV tasks  \cite{joint_2d_3d_MTL,li2022bevformer,bevformer_v2}.  We focus on multitask transformer models that address dense prediction tasks such as semantic segmentation, monocular depth estimation, surface normal prediction, saliency estimation, and object boundary detection. Those models typically require substantial numbers of queries but also have the potential for large efficiency gains by exploiting the grid structure of their features. Influential works with significant architectural innovations and strong results in that field include the following: In MulT \cite{Bhattacharjee_2022_CVPR} by Bhattacharjee \etal, the shared attention module for each task decoder receives its queries and keys from the corresponding shared representation in the model's encoder. MQTransformer \cite{Xu2022MultiTaskLW} demonstrated a method for more efficient attention computation across tasks by replacing per-pixel queries with fewer task-specific ones. A promising recent development in MTL for dense vision tasks is MTMamba \cite{lin2024mtmambaenhancingmultitaskdense} by Lin \etal. It is a State Space Model rather than a Transformer which provides better scaling behavior with the number of queries while still reaching a comparable level of prediction quality. Compared to Transformer-based models that achieve similar results, however, it uses significantly more learnable parameters and compute resources. 


\textbf{Deformable Mixer Transformer.} We implement our method based on the influential Multitask Vision Transformer architecture \textit{DeMT} introduced by Xu \etal in \cite{xyy2023DeMT} to demonstrate its impact on a well-established baseline. This model achieved state-of-the-art results with global attention across all tasks. It first uses a backbone pre-trained on Imagenet \cite{imagenet_paper} to extract a feature map from the input data in the image domain. This backbone can be chosen from several options including HRNetV2 \cite{8953615} and variants of the Swin transformer \cite{Swin}. Then, as part of the transformer encoder, the extracted features are passed on to multiple \textit{Deformable Mixer} modules, one per task, which refine it for each task. Thereafter, the output feature maps from these Deformable Mixers are fed into the decoder part of the model, the \textit{Task-aware Transformer Decoder}. This decoder consists of a \textit{Task Interaction Block} which allows information from one task to be propagated to the others. Finally, for each task, its feature map is further refined in its task-specific \textit{Task Query Block} consisting of a Multi-Headed Self-Attention (MHSA) module and a fully-connected Multi-Layer-Perceptron (MLP). 
The part of the architecture we focus on for this work is the Task Interaction Block. In DeMT, this block accepts the concatenation of the encoder outputs for all tasks as its set of input queries and performs MHSA on that, followed by an MLP. As outlined before, for multiple tasks, this Task Interaction Block exhibits poor scaling behavior and suboptimal prediction quality. Hence we replace it by our more efficient version described in \Cref{methodology}. 

\section{Methodology} \label{methodology}

\begin{figure}[h]
    \centering
    \includegraphics[width=0.5\linewidth]{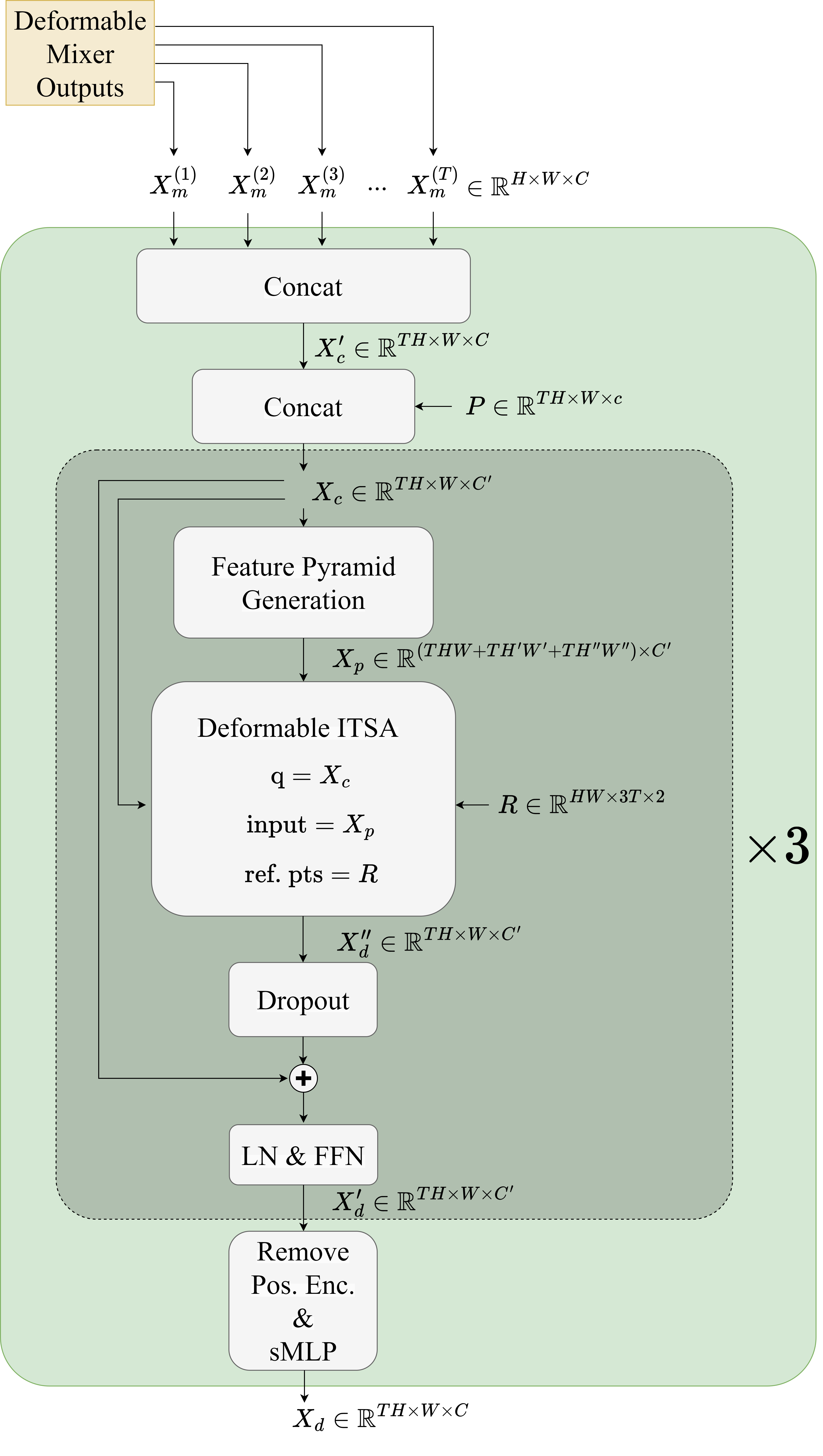}
    \caption{Schematic of our Task Interaction Block. LN refers to Layer Norm, FFN to a Feedforward Network and sMLP is a small MLP of a linear layer followed by a Layer Norm.}
    \label{task_interaction_block_figure}
\end{figure}

In the following, we describe our Deformable Inter-Task Self-Attention (ITSA) method in detail. It is a type of self-attention, i.e., its query set attends to itself. The main intuition behind our method is that for any query for a particular task, it allows the model to freely sample from a relatively small number of locations in all task's feature maps. Thus it can efficiently aggregate information from all tasks in response to a query from one task. This method is applicable to a wide range of MTL architectures also beyond CV, provided they use queries that attend to grid-structured features from different tasks. 

The core component of our Task Interaction Block for the DeMT model is Deformable ITSA. 
As shown in \Cref{task_interaction_block_figure}, our Task Interaction Block receives the concatenation of the output features from the previous Deformable Mixer modules for all tasks, as in DeMT. In the following, let $H$, $W$, and $C$ respectively denote the height, width, and number of channels of the full-scale feature maps for all tasks, and $T$ the number of tasks in the model. For the output feature from the Deformable Mixer belonging to task $t$, $X_m^{(t)} \in \mathbb{R}^{H \times W \times C}$, the Interaction Block accepts the concatenation along the height dimension:


\begin{figure*}[ht]
    \centering
    \includegraphics[width=\linewidth]{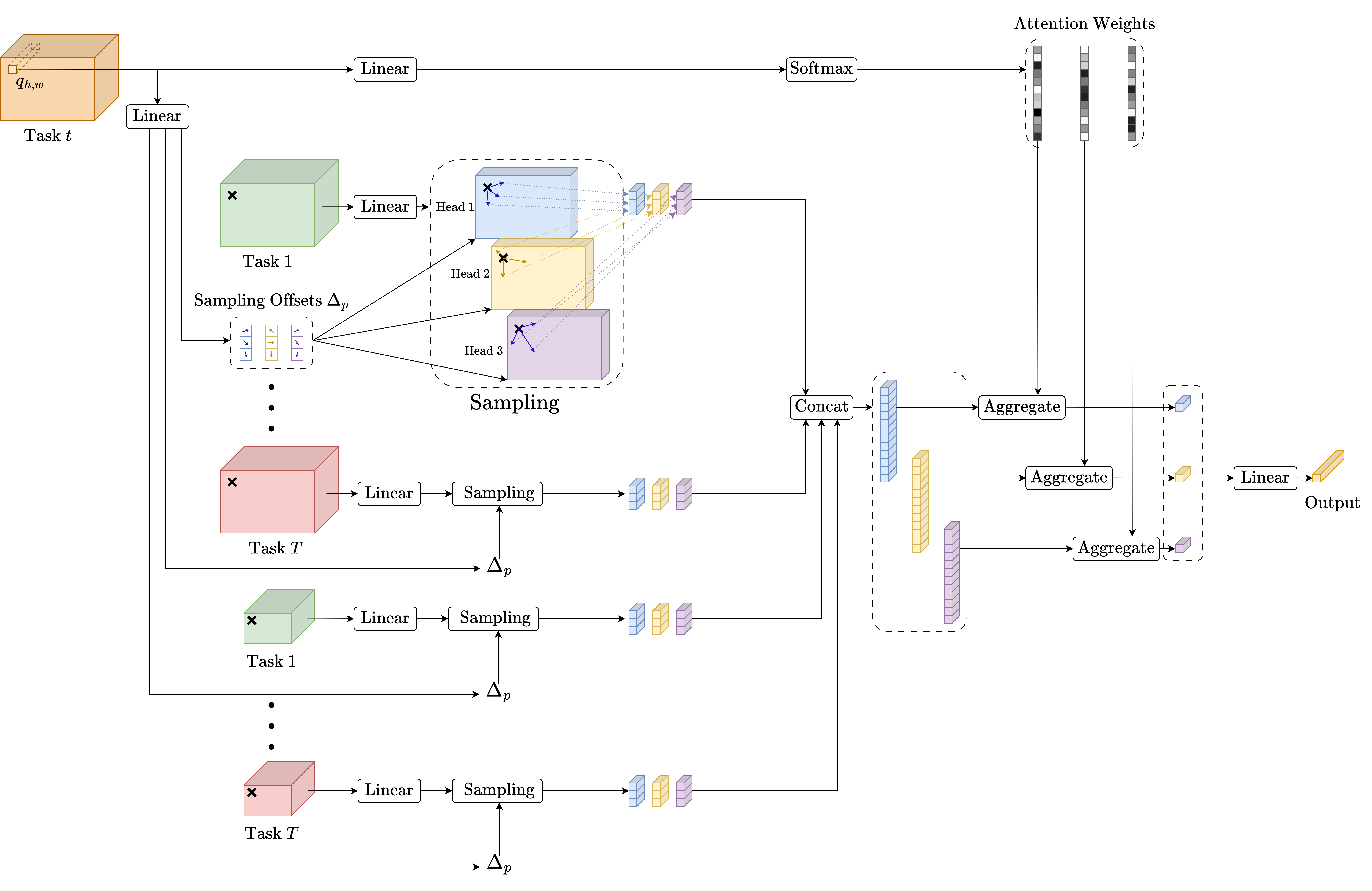}
    \caption{Illustration of our Deformable Inter-Task Self-Attention for one query $q_{h,w}$ from task $t$. The feature map for task $t$ on the left contains the query at position $(h, w)$, the feature pyramid (the green and red tensors) for all tasks is the input to the deformable attention, and the output on the right contains the refined features for query $q_{h,w}$ of task $t$ after ITSA. $\vectimes$ indicates the reference point $R'_{h,w}$ for query $q_{h,w}$. In this illustration, we only show two feature pyramid levels and three heads with three offsets each, the actual values for these are higher.}
    \label{ITSA_figure}
\end{figure*}

\begin{equation}
    X'_c = \text{Concat}(X_m^{(1)}, X_m^{(2)}, ..., X_m^{(T)}) \in \mathbb{R}^{TH \times W \times C}
\end{equation} 
In the next step, we concatenate a 2-dimensional sinusoidal positional embedding $P \in \mathbb{R}^{TH \times W \times c}$ of $c$ channels to that, yielding the feature 
\begin{equation}
    X_c = \text{Concat}(X'_c, P) \in \mathbb{R}^{TH \times W \times C'}
\end{equation}
$X_c$ is passed as the query feature to the Deformable ITSA and the input features are derived from it. 
For improved robustness against variable scales of patterns in the input data, the input features to Deformable ITSA are a feature pyramid, as can be seen in \Cref{ITSA_figure}.
We obtain these features as follows: For the full-scale feature map for task $t$, i.e., the portion of $X_c =: X_{p_0}$ that corresponds to task $t$, $X_{p_0}^{t}$, we compute two down-sampled versions for the smaller pyramid levels. To roughly halve the size of each spatial dimension in each down-sampling step, we use 3-by-3 convolutions with a stride of 2. 
The down-sampled maps are defined as follows: 

\begin{equation}
    X_{p_1}^t = \text{Conv1}(X_{p_0}^t) \in \mathbb{R}^{H'\times W' \times C'}
\end{equation}
for $t \in \{1, ..., T\}$,
\begin{equation}
    X_{p_1} = \text{Concat}(X_{p_1}^1, ..., X_{p_1}^T) \in \mathbb{R}^{TH'\times W' \times C'}
\end{equation}
and
\begin{equation}
    X_{p_2}^t = \text{Conv2}(X_{p_1}^t) \in \mathbb{R}^{H''\times W'' \times C'}
\end{equation}  
with $X_{p_2}$ derived from $[X_{p_2}^1, ..., X_{p_2}^T]$ analogously to $X_{p_1}$. From that, we obtain $X_p$ through flattening and concatenation:

\begin{equation}
    \begin{aligned}
        X_p &= \text{Concat}(\text{Flatten}(X_{p_l}) \text{ for } l \in \{0, 1, 2\}) \\
            &\in \mathbb{R}^{(THW+TH'W'+TH''W'')\times C'} 
    \end{aligned}
\end{equation}
We define the reference points for sampling input features for all queries as follows: For query $q_{h,w}$ at index $(h,w)$ for any task, its assigned reference point is 
\begin{equation}
    R'_{h,w}=\bigg(\frac{h+0.5}{H}, \frac{w+0.5}{W}\bigg),
\end{equation} 
i.e., the center of its cell in the query grid $X_c$.
For the down-sampled features, the reference points are the same as above since reference points are defined relative to the feature size, not in absolute coordinates. After flattening the height and width dimension of $R'$ into one and repeating it for all tasks and three pyramid levels, we obtain the tensor of reference points
\begin{equation}
    R \in \mathbb{R}^{HW\times3T\times2}
\end{equation}
\begin{table*}[t]
    \centering
    \caption{Comparison of our method with previous state-of-the-art approaches on the NYUD-v2 dataset. $\Delta\%$ denotes the improvement percentage over DeMT \cite{xyy2023DeMT}.}
    \begin{tabular}{l l c c c c c}
        \label{NYUD_results}
        \vspace{0cm}\\
        \hline \\
        \vspace{-0.7cm} \\
        Model   & Backbone  & \begin{tabular}{@{}c@{}}SemSeg\\(mIoU)$\uparrow$\end{tabular}  & \begin{tabular}{@{}c@{}}Depth\\(RMSE)$\downarrow$\end{tabular}     & \begin{tabular}{@{}c@{}}Normal\\(mErr)$\downarrow$\end{tabular}    & \begin{tabular}{@{}c@{}}Bound\\(odsF)$\uparrow$\end{tabular}     & $\Delta\%\uparrow$ \\
        \hline \\
        \vspace{-0.7cm} \\
        DeMT \cite{xyy2023DeMT}    & HRNet18   & 0.3760    & 0.6286    & 20.80     & \textbf{0.7718}    & 0.00    	\\
        Ours    & HRNet18   & \textbf{0.4039}    & \textbf{0.5819}   & \textbf{20.34} & 0.7712    & \textbf{4.25} \\  
        \hline\\
        \vspace{-0.7cm} \\
        single task baseline \cite{xyy2023DeMT} & Swin-T & 0.4292 & 0.6104 & 20.94 & 0.7622 & -3.42\\
        multi task baseline \cite{xyy2023DeMT} & Swin-T & 0.3878 & 0.6312 & 21.05 & 0.7560 & -6.87 \\
        MQTransformer \cite{Xu2022MultiTaskLW} & Swin-T & 0.4361 & 0.5979 & \textbf{20.05} & 0.7620 & -1.44\\
        DeMT \cite{xyy2023DeMT} & Swin-T    & 0.4636    & 0.5871    & 20.65 & 0.7690    & 0.00 \\
        Ours    & Swin-T    & \textbf{0.4806}    & \textbf{0.5567}    & 20.14 & \textbf{0.7769}    & \textbf{3.08}\\
        \hline\\
        \vspace{-0.7cm} \\
        single task baseline \cite{xyy2023DeMT} & Swin-S & 0.4892 & 0.5804 & 20.94 & 0.7720 & -4.20 \\
        multi task baseline \cite{xyy2023DeMT} & Swin-S & 0.4790 & 0.6053 & 21.17 & 0.7690 & -6.21 \\
        MQTransformer \cite{Xu2022MultiTaskLW} & Swin-S & 0.4918 & 0.5785 & 20.81 & 0.7700 & -3.89 \\
        DeMT \cite{xyy2023DeMT} & Swin-S & 0.5150 & 0.5474 & 20.02 & 0.7810 & 0.00\\
        Ours & Swin-S & \textbf{0.5259} & \textbf{0.5356} & \textbf{19.87} & \textbf{0.7859} & \textbf{1.42} \\
        \hline\\
        
    \end{tabular}
\end{table*}
\begin{table*}[t]
    \centering
    \caption{Comparison of our method with previous state-of-the-art approaches on the PASCAL-Context dataset. $\Delta\%$ denotes the improvement percentage over DeMT \cite{xyy2023DeMT}.}
    \begin{tabular}{l l c c c c c c}
        \label{PASCAL_results}
        \vspace{0cm}\\
        \hline \\
        \vspace{-0.7cm} \\
        Model   & Backbone  & \begin{tabular}{@{}c@{}}SemSeg\\(mIoU)$\uparrow$\end{tabular}  & \begin{tabular}{@{}c@{}}PartSeg\\(mIoU)$\uparrow$\end{tabular}     & \begin{tabular}{@{}c@{}}Sal\\(maxF)$\uparrow$\end{tabular}    & \begin{tabular}{@{}c@{}}Normal\\(mErr)$\downarrow$\end{tabular}  & \begin{tabular}{@{}c@{}}Bound\\(odsF)$\uparrow$\end{tabular}   & $\Delta\%\uparrow$ \\
        \hline \\
        \vspace{-0.7cm} \\
        DeMT \cite{xyy2023DeMT}    & HRNet18   & 0.5582    & \textbf{0.5708}    & 0.8417     & 14.21    & 0.7490    &	0.00\\
        Ours    & HRNet18   & \textbf{0.5821}    & 0.5706    & \textbf{0.8453} & \textbf{14.02}    & \textbf{0.7526} & \textbf{1.29}\\  
        \hline\\
        \vspace{-0.7cm} \\
        single task baseline \cite{xyy2023DeMT} & Swin-T & 0.6781 & 0.5632 & 0.8218 & 14.81 & 0.7090 & -1.38 \\
        multi task baseline \cite{xyy2023DeMT} & Swin-T & 0.6474 & 0.5325 & 0.7688 & 15.86 & 0.6900 & -6.60 \\
        MQTransformer \cite{Xu2022MultiTaskLW} & Swin-T & 0.6824 & 0.5705 & 0.8340 & 14.56 & 0.7110 & -0.31\\
        DeMT \cite{xyy2023DeMT} & Swin-T    & 0.6971    & 0.5718    & 0.8263 & 14.56    & 0.7120 & 0.00\\
        Ours    & Swin-T    & \textbf{0.6987}    & \textbf{0.5828}    & \textbf{0.8485} & \textbf{14.03}   & \textbf{0.7746} & \textbf{3.46}\\
        \hline\\
        
    \end{tabular}
\end{table*}
We pass $X_c$ as query feature, $X_p$ as input feature and the reference points $R$ to the Deformable Attention function as defined in \cite{zhu2021deformable}. 
With these input tensors, we refer to performing Deformable Attention on them as Deformable Inter-Task Self-Attention. 
As output we receive the features $X''_d$, i.e., the refined task feature maps after attending to all task's features in Deformable ITSA. \Cref{ITSA_figure} shows how Deformable ITSA is computed for one query $q_{h,w}$ from task $t$. For each reference point and task feature map, a fixed number of offsets $\Delta_p$ are computed from the query feature through a linear layer. These define the positions relative to the reference point that the input is being sampled at. Since the offsets and reference points are generally not integral, bilinear interpolation is applied to generate a sample. Analogously to the offset, based on the query, attention weights are generated using a linear layer followed by a softmax. Similarly to Multi-Head Attention \cite{attention_is_all_you_need}, the input feature maps in $X_p$ are not sampled directly, but instead multiple learned projections of them are, one for each head. The attention weights are then used to aggregate the samples for the different heads, tasks, and feature pyramid levels to generate the output of Deformable ITSA for a query. 
Applying this operation to all queries in $X_c$, yields the following output:
\begin{equation}
    X''_d = \text{Def.ITSA}(X_c, X_p, R) \in \mathbb{R}^{TH \times W \times C'}  
\end{equation}
For regularization we feed $X''_d$ through a Dropout layer \cite{dropout_paper} with a drop probability of 0.1 and for better preservation of high-frequency features, we add $X_c$ to it via  a residual connection. Thereafter, we apply Layer Norm (LN) \cite{Ba2016LayerN} and a small Feedforward Network (FFN) to generate
\begin{equation}
    \begin{aligned}
            X'_d    &= \text{FFN}(\text{LN}(X_c + \text{Dropout}(X''_d)))\\
                    &\in \mathbb{R}^{TH \times W \times C'}  
    \end{aligned}
\end{equation}
In total, we perform three ITSA refinement steps, i.e., we update $X_c := X'_d$ and repeat the feature pyramid computation, the Deformable ITSA, Dropout, LN, and FFN two more times to arrive at the final $X'_d$. From that, we remove the last $c$ channels containing the positional information and, following \cite{xyy2023DeMT}, apply a small MLP of a linear layer followed by a Layer Norm to generate $X_d \in \mathbb{R}^{TH \times W \times C}$ as the final output from our Task Interaction Block. 

\begin{table*}[t]
    \centering
    \caption{Ablation on components of our Task Interaction Block evaluated on NYUD-v2 dataset with HRNet18 backbone. \textit{Deformable ITSA} replaces the Task Interaction Block with our version. \textit{3-by-3 conv. downsample} refers to performing the downsampling with a 3-by-3 convolution with a stride of 2 instead of 1-by-1 convolution followed by 2-by-2 max-pooling. \textit{3 refinement steps} refers to performing 3 Deformable ITSA steps instead of 1.}
    \begin{tabular}{l c c c c c}
        \label{component_ablation}
        \vspace{0cm}\\
        \hline \\
        \vspace{-0.7cm} \\
        Model  & \begin{tabular}{@{}c@{}}SemSeg\\(IoU)$\uparrow$\end{tabular}  & \begin{tabular}{@{}c@{}}Depth\\(RMSE)$\downarrow$\end{tabular}     & \begin{tabular}{@{}c@{}}Normal\\(mErr)$\downarrow$\end{tabular}    & \begin{tabular}{@{}c@{}}Bound\\(odsF)$\uparrow$\end{tabular}     & $\Delta\%\uparrow$ \\
        \hline \\
        \vspace{-0.7cm} \\
        DeMT baseline \cite{xyy2023DeMT} & 0.3760 & 0.6286 & 20.80 & 0.7718 & 0.00 \\
        Deformable ITSA &  0.3907 & 0.5877 & 20.41 & 0.7721 & 3.08\\
        + 3-by-3 conv. downsample & 0.3958 & 0.5954 & 20.34 & 0.7737 & 3.25\\
        + positional encoding & 0.4004 & 0.5943 & 20.30 & 0.7732 & 3.63\\
        + 3 refinement steps & 0.4039 & 0.5819 & 20.34 & 0.7712 & 4.25\\
        \hline\\
        
    \end{tabular}
\end{table*}

\begin{table*}[t]
    \centering
    \caption{Ablation on gradient scaling factors $\lambda$ evaluated on NYUD-v2 dataset with HRNet18 backbone. }
    \begin{tabular}{l c c c c c}
        \label{gradient_scale_ablation}
        \vspace{0cm}\\
        \hline \\
        \vspace{-0.7cm} \\
        $\lambda$  & \begin{tabular}{@{}c@{}}SemSeg\\(IoU)$\uparrow$\end{tabular}  & \begin{tabular}{@{}c@{}}Depth\\(RMSE)$\downarrow$\end{tabular}     & \begin{tabular}{@{}c@{}}Normal\\(mErr)$\downarrow$\end{tabular}    & \begin{tabular}{@{}c@{}}Bound\\(odsF)$\uparrow$\end{tabular}     & $\Delta\%\uparrow$ \\
        \hline \\
        \vspace{-0.7cm} \\
        $1\times$ &     0.3989 & 0.6010 & 20.35 & 0.7743 & 0.00\\
        $10 \times$ &   0.4020 & 0.5939 & 20.28 & 0.7737 & 0.56\\
        $30 \times$ &   0.4031 & 0.5920 & 20.31 & 0.7730 & 0.65\\
        $50 \times$ &   0.4088 & 0.5915 & 20.32 & 0.7729 & 1.01\\
        $100 \times$ &  0.4039 & 0.5819 & 20.34 & 0.7712 & 1.02\\
        $200 \times$ &  0.0199 & 1.5089 & 45.34 & 0.4169 & -104\\
        \hline\\
        
    \end{tabular}
\end{table*}

\begin{table*}[t]
    \centering
    \caption{Ablation on the number of feature pyramid levels evaluated on NYUD-v2 dataset with HRNet18 backbone. }
    \begin{tabular}{l c c c c c}
        \label{feature_pyramid_ablation}
        \vspace{0cm}\\
        \hline \\
        \vspace{-0.7cm} \\
        Pyramid Levels  & \begin{tabular}{@{}c@{}}SemSeg\\(IoU)$\uparrow$\end{tabular}  & \begin{tabular}{@{}c@{}}Depth\\(RMSE)$\downarrow$\end{tabular}     & \begin{tabular}{@{}c@{}}Normal\\(mErr)$\downarrow$\end{tabular}    & \begin{tabular}{@{}c@{}}Bound\\(odsF)$\uparrow$\end{tabular}     & $\Delta\%\uparrow$ \\
        \hline \\
        \vspace{-0.7cm} \\
        1 &  0.3979 & 0.5901 & 20.42 & 0.7734 & 0.00\\
        2 &  0.4028 & 0.5853 & 20.43 & 0.7707 & 0.41\\
        3 &  0.4039 & 0.5819 & 20.34 & 0.7712 & 0.75\\
        \hline\\
        
    \end{tabular}
\end{table*}

\begin{figure*}[t]
    \centering
    \addtolength{\tabcolsep}{-0.15cm}
    \begin{tabular}{ccccc}
        Input Image & Sem. Seg. & Depth & Normal & Boundary \\
    
        \includegraphics[width=0.19\textwidth]{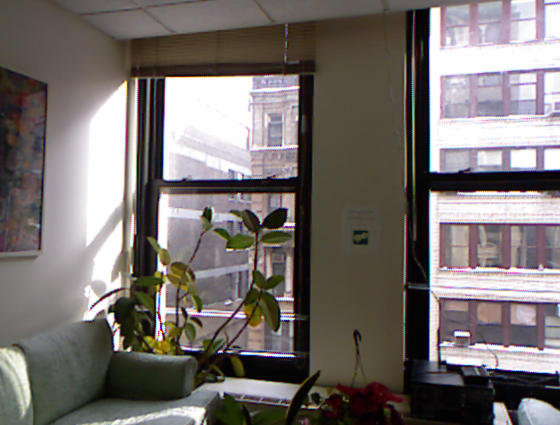} \hspace{0.015\textwidth} &
        \includegraphics[width=0.19\textwidth]{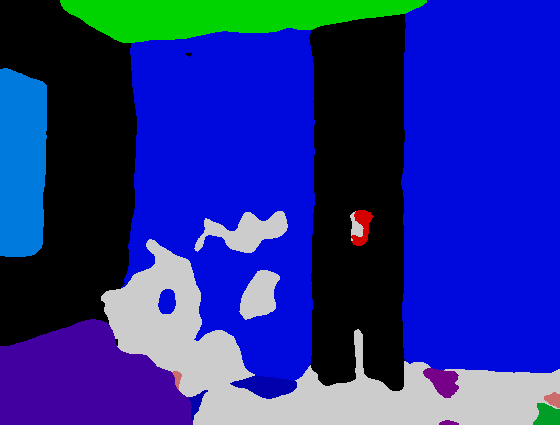} \hspace{0.015\textwidth}  & 
        \includegraphics[width=0.19\textwidth]{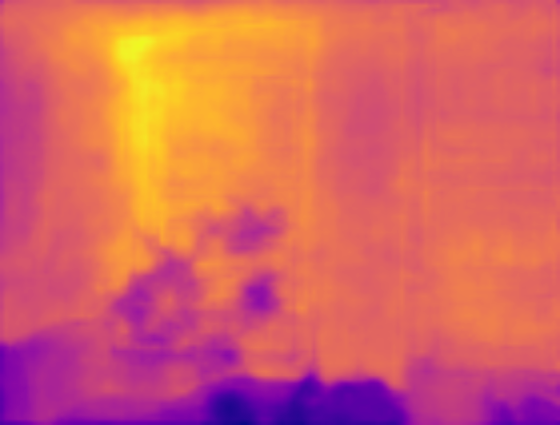} \hspace{0.015\textwidth}  & 
        \includegraphics[width=0.19\textwidth]{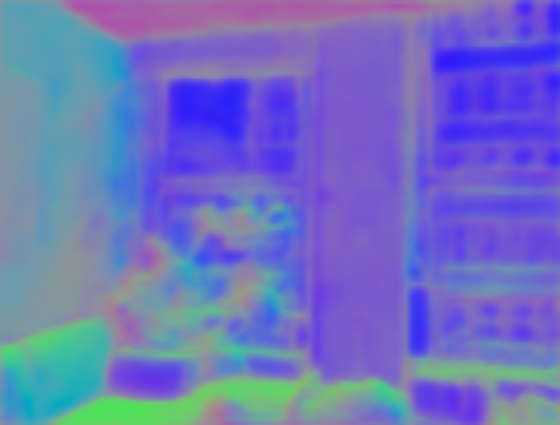} \hspace{0.015\textwidth}  & 
        \includegraphics[width=0.19\textwidth]{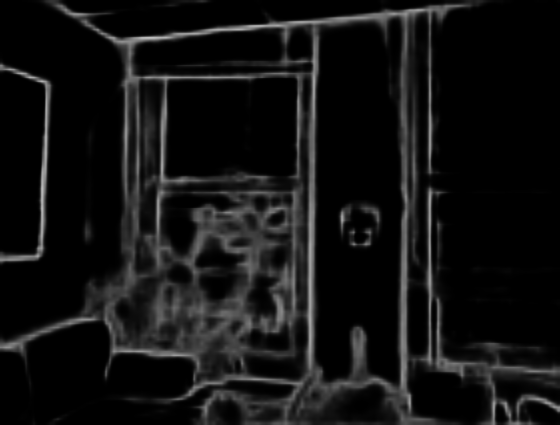} \\
        \vspace{-0.45cm} \\

        \includegraphics[width=0.19\textwidth]{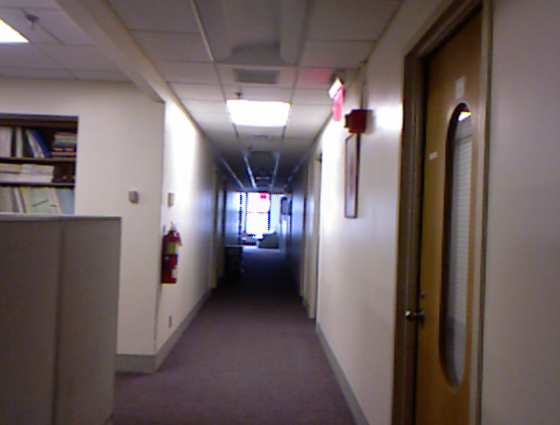} \hspace{0.015\textwidth}  &
        \includegraphics[width=0.19\textwidth]{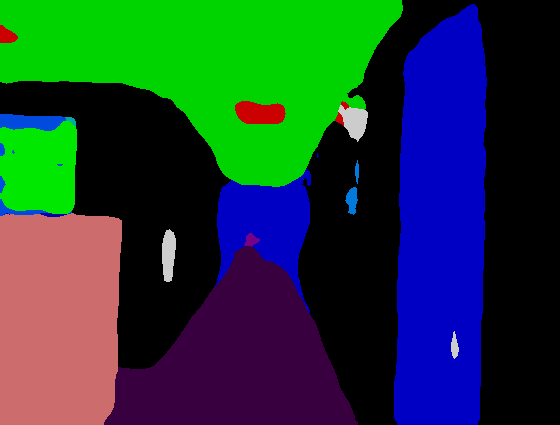} \hspace{0.015\textwidth}  & 
        \includegraphics[width=0.19\textwidth]{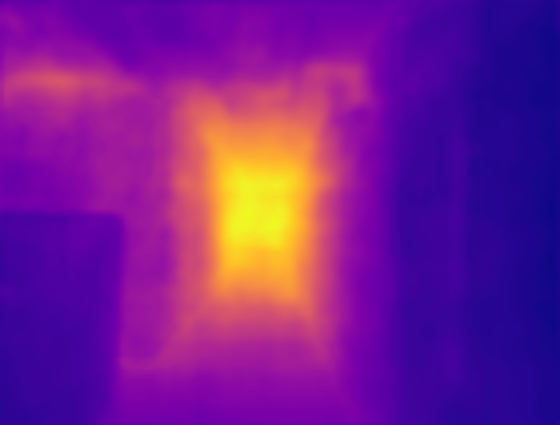} \hspace{0.015\textwidth}   & 
        \includegraphics[width=0.19\textwidth]{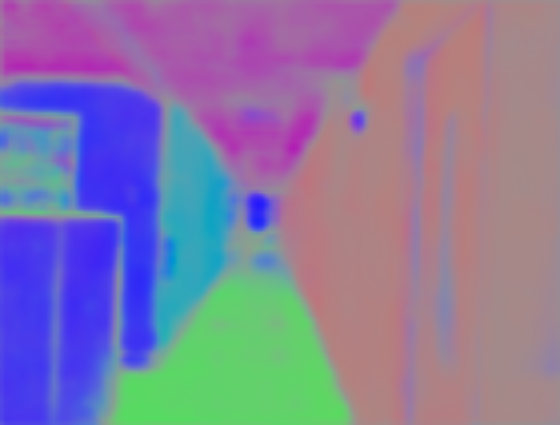} \hspace{0.015\textwidth}  & 
        \includegraphics[width=0.19\textwidth]{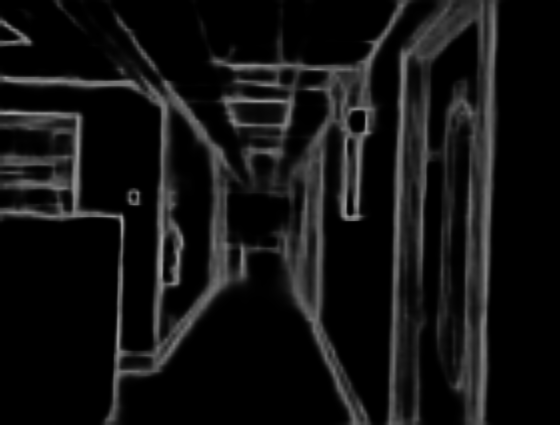} \\
        \vspace{-0.45cm} \\

        \includegraphics[width=0.19\textwidth]{qualitative_examples/results_sample_76/0076.png} \hspace{0.015\textwidth}  &
        \includegraphics[width=0.19\textwidth]{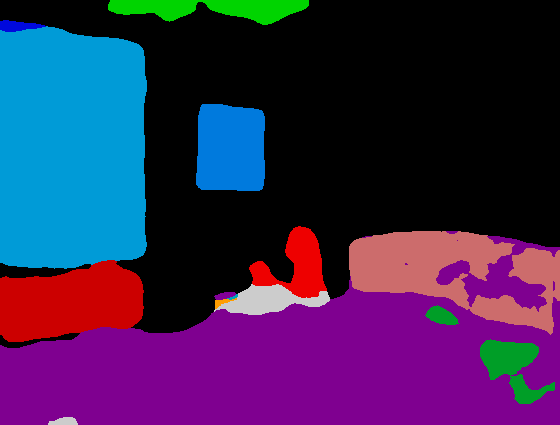} \hspace{0.015\textwidth}  & 
        \includegraphics[width=0.19\textwidth]{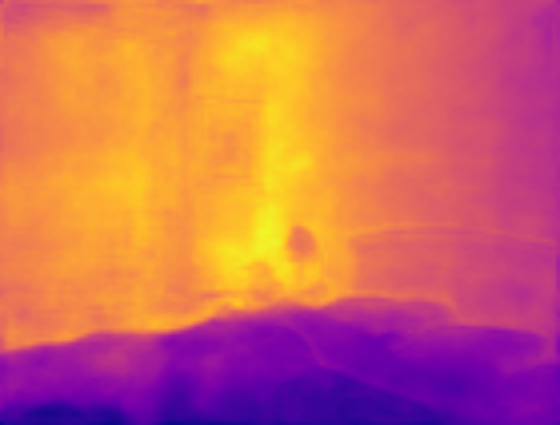} \hspace{0.015\textwidth}  & 
        \includegraphics[width=0.19\textwidth]{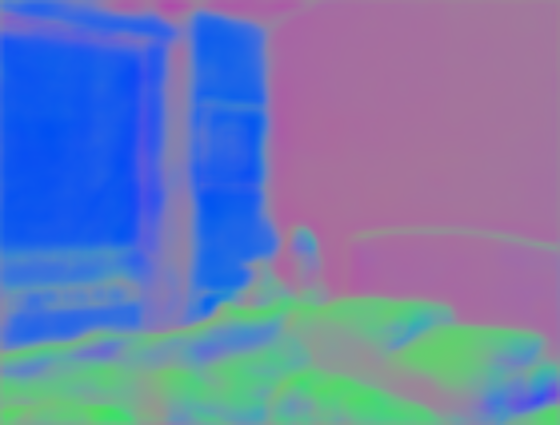} \hspace{0.015\textwidth}  & 
        \includegraphics[width=0.19\textwidth]{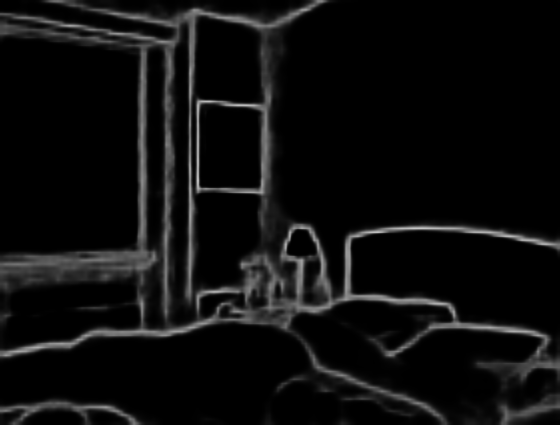} \\
    \end{tabular}
    \caption{Qualitative examples from the NYUD-v2 validation set produced with our model.}
\end{figure*}
    
\section{Experiments}
To validate our method and demonstrate its consistent effectiveness across multiple backbone architectures, model sizes, and datasets, we conduct experiments on three different backbones and two prominent multi-task vision datasets.

\subsection{Experimental Setup}
\textbf{Implementation Details.} 
We evaluate our Deformable ITSA method integrated into the DeMT \cite{xyy2023DeMT} architecture, as described before.  To ensure comparability with DeMT, we train the model using the SGD optimizer with the same values for the learning rate of $10^{-3}$ and the weight decay of $5\cdot10^{-4}$. Our loss function is a weighted sum of the individual task losses. For comparability with previous work, we use the same task loss functions and loss weights from DeMT without modification. As training hardware, we used one A100 80GB GPU for each experiment.

\textbf{Model Hyperparameters.}
The inputs and outputs of our Task Interaction Block all have a channel number $C=256$, thus maintaining compatibility with the DeMT architecture. For the Deformable ITSA module, a positional encoding of $c=24$ channels is concatenated, leading to an overall channel count of $C'=280$. In our grid search, we found that using more channels for the positional embedding did not yield additional improvements. Within the Deformable ITSA block, we use 4 heads and sample 16 offsets around every reference point. During our initial experiments, we observed that the learned offsets were very close to their initializations. Hence, as a new hyperparameter, we introduced a scaling factor $\lambda$ for the gradient of the Deformable ITSA module that allows the model to learn meaningful sampling offsets and improved its performance significantly. We found that a factor of 100 performed best in our experiments. In \Cref{gradient_scale_ablation}, an ablation on $\lambda$ can be seen, illustrating the impact of different choices for it.

\textbf{Datasets.}
We evaluate our method on the public NYUD-v2 \cite{NYUD_dataset_paper} and PASCAL-Context \cite{PASCAL_dataset_paper} datasets. In the multi-task computer vision literature, these are well established as benchmark datasets. 
NYUD-v2 contains 795 images for training and 654 images for validation. Each of those images has a resolution of $560\times425$ pixels and depicts an indoor scene. These images are densely annotated with their semantic segmentation, depth information, surface normals, and object boundaries. 
The PASCAL-Context dataset is split into 4998 training and 5105 validation images. Its images depict a variety of both indoor and outdoor scenes. Each image has dense pixel-wise labels to train semantic segmentation, human parts segmentation, saliency estimation, surface normal prediction, and object boundary estimation. 
For the HRNet18 backbone, our results are not directly comparable to the ones from the DeMT paper as the authors did not publish the necessary configuration files for that setup. Hence, for that backbone, we had to create our own files and train both the DeMT baseline and our own model with those. 

\textbf{Backbones.}
Our method is evaluated both on the CNN-based HRNet-V2-W18-small (HRNet18) backbone (3.9 M parameters) \cite{8953615}, as well as the Vision Transformer backbones Swin-T (27.5 M) and Swin-S (48.8 M) \cite{Swin}.  

\textbf{Metrics.}
Across both datasets, we evaluate six tasks in total. For these, their respective performance metrics are the following: For both semantic segmentation and human parts segmentation, we report the \textit{mean Intersection over Union (mIoU)} score, performance on the depth estimation task is measured by the \textit{Root Mean Squared Error (RMSE)}, surface normal estimation quality is judged by the \textit{mean Error (mErr)}, for the boundary estimation task we report the \textit{optimal dataset-scale F-measure (odsF)}, and the quality of the saliency estimation is quantified by the \textit{maximum F-measure (maxF)}. For all tasks, these are the standard metrics commonly reported in the relevant literature. We additionally provide the \textit{average improvement percentage ($\Delta \%$)} across all tasks relative to DeMT which is our baseline. 

\subsection{Results}
\textbf{NYUD-v2 dataset.}
In \Cref{NYUD_results}, we show the results on the NYUD-v2 dataset. For all backbones, our method achieves substantially enhanced results compared to the DeMT baseline. Assessing the average improvements, it is evident that the quality increases are more pronounced for smaller backbones. This is to be expected as for stronger features from the larger backbones, the metrics are already significantly better, thus making large improvements more challenging. Especially for the smaller HRNet-18 backbone, we observe outstanding improvements: There, we achieve $7.4\%$ better scores for both semantic segmentation and depth estimation while still improving the other two tasks substantially. Also for the Swin-T backbone, we see a performance increase of $5.2\%$ for the depth estimation task while all other tasks are also improved by at least $1\%$. 

\textbf{PASCAL-Context dataset.}
\Cref{PASCAL_results} shows the results on the PASCAL-Context dataset. Apart from the Human Parts Segmentation task with the HRNet18 backbone, we observe better performance in every task metric using our Task Interaction Block compared to the DeMT baseline. For both backbones evaluated, we observe substantial average improvements across the tasks by $1.29\%$ and $3.46\%$, respectively.  

\subsection{Runtime Metrics}
To illustrate the significant improvement in required compute resources with our method, we calculated the required FLOPs and measured the latency for the attention module in the Task Interaction Block during inference. We compare both of these metrics against those achieved with DeMT \cite{xyy2023DeMT}. For both versions, we measured inference time on the same hardware, an RTX A3000 GPU with 6GB of GPU memory. For both experiments, we evaluated with an HRNet18 backbone on the NYUD-v2 dataset with a batch size of 1, 3 feature pyramid levels, and a single refinement layer. 

\begin{table}[ht]
    \centering
    \caption{Comparison of compute resource requirements for attention computation between the DeMT baseline and our method.}
    \begin{tabular}{l c c}
        \label{runtime_metrics_table}
        \vspace{0cm}\\
        \hline \\
        \vspace{-0.7cm} \\
        Model  & FLOPs $\downarrow$ & Inference Latency $\downarrow$\\
        \hline \\
        \vspace{-0.7cm} \\
        DeMT \cite{xyy2023DeMT} &  3.75T & 10.72s\\
        Ours &  0.24T & 1.34s\\
        \hline\\
        
    \end{tabular}
\end{table}

As can be seen in \Cref{runtime_metrics_table}, our method reduces the amount of resources required to compute the attention by roughly an order of magnitude. 

\subsection{Ablations}
\textbf{Task Interaction Block.}
To demonstrate the impact of the individual features of our implementation, in \Cref{component_ablation} the ablation on components of our Task Interaction Block can be seen. From the improvement percentages compared to the DeMT baseline, it is evident that replacing global MHSA with a stripped-down version of our Deformable ITSA (in the second row) already leads to a large relative improvement in the average of all task metrics. This version only performs a single Deformable ITSA step, has no positional encoding appended to $X_c$, and performs the downsampling for the feature pyramid generation using a 1-by-1 convolution followed by 2-by-2 max-pooling. Applying a 3-by-3 convolution with stride 2 to downsample, appending a positional encoding, and using 3 refinement steps all have smaller relative impacts. However, their smaller individual contributions still add up to a substantial improvement for the fully-featured model in comparison to the baseline. 

\textbf{Gradient Scaling Factors.}
\Cref{gradient_scale_ablation} shows the ablation on gradient scaling factors $\lambda$ used in the Deformable ITSA, underscoring the importance of choosing an appropriate value for that hyperparameter. Comparing the average improvement percentages to the version with factor 1 that does not scale the gradient at all, we observe that factors of 10 and 30 already lead to improved metrics, but scaling by 50 or 100 appears to be close to the optimum. To illustrate that scaling beyond our choice of 100 yields no significant further improvement, we also experimented with a factor of 200, at which point the model's metrics collapse. This is due to that fact that for such a large scaling factor, a substantial fraction of sampling points would lie outside the feature map, necessitating clipping them to locations around the map's edge. That places the model in a deep local minimum where for most queries and offsets only those edge locations can be sampled, explaining the poor result.

\textbf{Feature Pyramid Levels.}
\Cref{feature_pyramid_ablation} demonstrates that additional pyramid levels improve Semantic Segmentation the most, with a smaller impact on the Depth Estimation task. Since the performance for the remaining two tasks roughly remains constant, the overall average improvement is not particularly large. We would still argue, however, that keeping multiple feature pyramid levels in the model is worthwhile for the improved segmentation performance alone.  


\section{Conclusion}
In this work, we introduced our Deformable Inter-Task Self-Attention, a method to significantly reduce the compute costs associated with attending between tasks in Vision Transformer models with many queries. We demonstrated that our approach leads to improvements in both FLOPs count and Inference Latency by roughly a factor of 10. At the same time we were still able to improve the prediction quality of the tasks in the model by a considerable amount across different backbones and datasets. 

A limitation of our method at the moment is its focus on dense prediction tasks. In future work, we intend to generalize our approach to other, less structured tasks, as well. In particular, we would like to extend our method to support the task of object detection, too.  This would involve having to learn to compute the reference points from the object queries, as we could no longer just use the position of a query in the feature map as its reference point.

\bibliographystyle{splncs04}
\bibliography{main}

\end{document}